%% file: main.tex
\documentclass{article}

\PassOptionsToPackage{numbers, compress}{natbib}
\usepackage{graphicx}
 \usepackage{amsmath}
 \usepackage{enumitem}
\usepackage{bm}
\usepackage{multirow}
 \usepackage[creativeai,final]{neurips_2025}

\usepackage[utf8]{inputenc} 
\usepackage[T1]{fontenc}    
\usepackage{hyperref}       
\usepackage{url}            
\usepackage{booktabs}       
\usepackage{amsfonts}       
\usepackage{nicefrac}       
\usepackage{microtype}      
\usepackage{xcolor}         

\title{Knolling Bot: Teaching Robots the Human Notion of Tidiness}

\author{%
  Yuhang Hu$^{1}$,\And
  Judah Goldfeder$^{1}$,\And
  Zhizhuo Zhang$^{1}$,\And
  Xinyue Zhu$^{1}$,\And
  Ruibo Liu$^{1}$,\And
  Philippe M.~Wyder$^{1}$,\And
  Jiong Lin$^{1}$,\And
  Hod Lipson$^{1}$
  \AND
  $^{1}$Columbia University \\
  \\
  https://www.youtube.com/watch?v=jCxykR4iP0I
}

\begin{document}

\maketitle

\input{sec/0_abstract}    
\input{sec/1_main}

\bibliography{main}
\bibliographystyle{plainnat}
\appendix

\input{sec/X_suppl}


\newpage
\section*{NeurIPS Paper Checklist}

The checklist is designed to encourage best practices for responsible machine learning research, addressing issues of reproducibility, transparency, research ethics, and societal impact. Do not remove the checklist: {\bf The papers not including the checklist will be desk rejected.} The checklist should follow the references and follow the (optional) supplemental material.  The checklist does NOT count towards the page
limit. 

Please read the checklist guidelines carefully for information on how to answer these questions. For each question in the checklist:
\begin{itemize}
    \item You should answer \answerYes{}, \answerNo{}, or \answerNA{}.
    \item \answerNA{} means either that the question is Not Applicable for that particular paper or the relevant information is Not Available.
    \item Please provide a short (1–2 sentence) justification right after your answer (even for NA). 
\end{itemize}

{\bf The checklist answers are an integral part of your paper submission.} They are visible to the reviewers, area chairs, senior area chairs, and ethics reviewers. You will be asked to also include it (after eventual revisions) with the final version of your paper, and its final version will be published with the paper.

The reviewers of your paper will be asked to use the checklist as one of the factors in their evaluation. While "\answerYes{}" is generally preferable to "\answerNo{}", it is perfectly acceptable to answer "\answerNo{}" provided a proper justification is given (e.g., "error bars are not reported because it would be too computationally expensive" or "we were unable to find the license for the dataset we used"). In general, answering "\answerNo{}" or "\answerNA{}" is not grounds for rejection. While the questions are phrased in a binary way, we acknowledge that the true answer is often more nuanced, so please just use your best judgment and write a justification to elaborate. All supporting evidence can appear either in the main paper or the supplemental material, provided in appendix. If you answer \answerYes{} to a question, in the justification please point to the section(s) where related material for the question can be found.

IMPORTANT, please:
\begin{itemize}
    \item {\bf Delete this instruction block, but keep the section heading ``NeurIPS Paper Checklist"},
    \item  {\bf Keep the checklist subsection headings, questions/answers and guidelines below.}
    \item {\bf Do not modify the questions and only use the provided macros for your answers}.
\end{itemize}


\begin{enumerate}

\item {\bf Claims}
    \item[] Question: Do the main claims made in the abstract and introduction accurately reflect the paper's contributions and scope?
    \item[] Answer: \answerYes{} 
    \item[] Justification:  In the abstract and introduction, we claim the contributions, including our proposed self-supervised framework, use of transformers with GMMs for multi-target placement, and deployment on real robots. 
    \item[] Guidelines:
    \begin{itemize}
        \item The answer NA means that the abstract and introduction do not include the claims made in the paper.
        \item The abstract and/or introduction should clearly state the claims made, including the contributions made in the paper and important assumptions and limitations. A No or NA answer to this question will not be perceived well by the reviewers. 
        \item The claims made should match theoretical and experimental results, and reflect how much the results can be expected to generalize to other settings. 
        \item It is fine to include aspirational goals as motivation as long as it is clear that these goals are not attained by the paper. 
    \end{itemize}

\item {\bf Limitations}
    \item[] Question: Does the paper discuss the limitations of the work performed by the authors?
    \item[] Answer: \answerYes{} 
    \item[] Justification: The limitations are discussed in the conclusion and within the experimental analysis, noting issues such as the scope of generalization to 3D environments and the difficulty of capturing subjective preferences. 
    \item[] Guidelines:
    \begin{itemize}
        \item The answer NA means that the paper has no limitation while the answer No means that the paper has limitations, but those are not discussed in the paper. 
        \item The authors are encouraged to create a separate "Limitations" section in their paper.
        \item The paper should point out any strong assumptions and how robust the results are to violations of these assumptions (e.g., independence assumptions, noiseless settings, model well-specification, asymptotic approximations only holding locally). The authors should reflect on how these assumptions might be violated in practice and what the implications would be.
        \item The authors should reflect on the scope of the claims made, e.g., if the approach was only tested on a few datasets or with a few runs. In general, empirical results often depend on implicit assumptions, which should be articulated.
        \item The authors should reflect on the factors that influence the performance of the approach. For example, a facial recognition algorithm may perform poorly when image resolution is low or images are taken in low lighting. Or a speech-to-text system might not be used reliably to provide closed captions for online lectures because it fails to handle technical jargon.
        \item The authors should discuss the computational efficiency of the proposed algorithms and how they scale with dataset size.
        \item If applicable, the authors should discuss possible limitations of their approach to address problems of privacy and fairness.
        \item While the authors might fear that complete honesty about limitations might be used by reviewers as grounds for rejection, a worse outcome might be that reviewers discover limitations that aren't acknowledged in the paper. The authors should use their best judgment and recognize that individual actions in favor of transparency play an important role in developing norms that preserve the integrity of the community. Reviewers will be specifically instructed to not penalize honesty concerning limitations.
    \end{itemize}

\item {\bf Theory assumptions and proofs}
    \item[] Question: For each theoretical result, does the paper provide the full set of assumptions and a complete (and correct) proof?
    \item[] Answer: \answerNA{} 
    \item[] Justification: The paper does not include formal theoretical proofs or theorems; it focuses on an empirical self-supervised learning system.
    \item[] Guidelines:
    \begin{itemize}
        \item The answer NA means that the paper does not include theoretical results. 
        \item All the theorems, formulas, and proofs in the paper should be numbered and cross-referenced.
        \item All assumptions should be clearly stated or referenced in the statement of any theorems.
        \item The proofs can either appear in the main paper or the supplemental material, but if they appear in the supplemental material, the authors are encouraged to provide a short proof sketch to provide intuition. 
        \item Inversely, any informal proof provided in the core of the paper should be complemented by formal proofs provided in appendix or supplemental material.
        \item Theorems and Lemmas that the proof relies upon should be properly referenced. 
    \end{itemize}

    \item {\bf Experimental result reproducibility}
    \item[] Question: Does the paper fully disclose all the information needed to reproduce the main experimental results of the paper to the extent that it affects the main claims and/or conclusions of the paper (regardless of whether the code and data are provided or not)?
    \item[] Answer: \answerYes{} 
    \item[] Justification:  We provide a complete description of the architecture, dataset generation, loss functions, and training procedures in Section 4 and supplementary materials.
    \item[] Guidelines:
    \begin{itemize}
        \item The answer NA means that the paper does not include experiments.
        \item If the paper includes experiments, a No answer to this question will not be perceived well by the reviewers: Making the paper reproducible is important, regardless of whether the code and data are provided or not.
        \item If the contribution is a dataset and/or model, the authors should describe the steps taken to make their results reproducible or verifiable. 
        \item Depending on the contribution, reproducibility can be accomplished in various ways. For example, if the contribution is a novel architecture, describing the architecture fully might suffice, or if the contribution is a specific model and empirical evaluation, it may be necessary to either make it possible for others to replicate the model with the same dataset, or provide access to the model. In general. releasing code and data is often one good way to accomplish this, but reproducibility can also be provided via detailed instructions for how to replicate the results, access to a hosted model (e.g., in the case of a large language model), releasing of a model checkpoint, or other means that are appropriate to the research performed.
        \item While NeurIPS does not require releasing code, the conference does require all submissions to provide some reasonable avenue for reproducibility, which may depend on the nature of the contribution. For example
        \begin{enumerate}
            \item If the contribution is primarily a new algorithm, the paper should make it clear how to reproduce that algorithm.
            \item If the contribution is primarily a new model architecture, the paper should describe the architecture clearly and fully.
            \item If the contribution is a new model (e.g., a large language model), then there should either be a way to access this model for reproducing the results or a way to reproduce the model (e.g., with an open-source dataset or instructions for how to construct the dataset).
            \item We recognize that reproducibility may be tricky in some cases, in which case authors are welcome to describe the particular way they provide for reproducibility. In the case of closed-source models, it may be that access to the model is limited in some way (e.g., to registered users), but it should be possible for other researchers to have some path to reproducing or verifying the results.
        \end{enumerate}
    \end{itemize}

\item {\bf Open access to data and code}
    \item[] Question: Does the paper provide open access to the data and code, with sufficient instructions to faithfully reproduce the main experimental results, as described in supplemental material?
    \item[] Answer: \answerYes{}
    \item[] Justification: We will release our code and the full dataset of tidy arrangements with documentation upon publication. The GitHub repo is ready: \url{https://github.com/H-Y-H-Y-H/knolling_bot}
    \item[] Guidelines:
    \begin{itemize}
        \item The answer NA means that paper does not include experiments requiring code.
        \item Please see the NeurIPS code and data submission guidelines (\url{https://nips.cc/public/guides/CodeSubmissionPolicy}) for more details.
        \item While we encourage the release of code and data, we understand that this might not be possible, so “No” is an acceptable answer. Papers cannot be rejected simply for not including code, unless this is central to the contribution (e.g., for a new open-source benchmark).
        \item The instructions should contain the exact command and environment needed to run to reproduce the results. See the NeurIPS code and data submission guidelines (\url{https://nips.cc/public/guides/CodeSubmissionPolicy}) for more details.
        \item The authors should provide instructions on data access and preparation, including how to access the raw data, preprocessed data, intermediate data, and generated data, etc.
        \item The authors should provide scripts to reproduce all experimental results for the new proposed method and baselines. If only a subset of experiments are reproducible, they should state which ones are omitted from the script and why.
        \item At submission time, to preserve anonymity, the authors should release anonymized versions (if applicable).
        \item Providing as much information as possible in supplemental material (appended to the paper) is recommended, but including URLs to data and code is permitted.
    \end{itemize}

\item {\bf Experimental setting/details}
    \item[] Question: Does the paper specify all the training and test details (e.g., data splits, hyperparameters, how they were chosen, type of optimizer, etc.) necessary to understand the results?
    \item[] Answer: \answerYes{}
    \item[] Justification: We describe the training and test splits, architecture, loss weights, optimizer, and domain randomization settings in code and technical details in the appendix.
    \item[] Guidelines:
    \begin{itemize}
        \item The answer NA means that the paper does not include experiments.
        \item The experimental setting should be presented in the core of the paper to a level of detail that is necessary to appreciate the results and make sense of them.
        \item The full details can be provided either with the code, in appendix, or as supplemental material.
    \end{itemize}

\item {\bf Experiment statistical significance}
    \item[] Question: Does the paper report error bars suitably and correctly defined or other appropriate information about the statistical significance of the experiments?
    \item[] Answer: \answerYes{} 
    \item[] Justification: We report standard deviations in all quantitative tables and discuss variance due to object count and architecture. See Table \ref{tab:sim_eval} and \ref{tab:ablation}
    \item[] Guidelines:
    \begin{itemize}
        \item The answer NA means that the paper does not include experiments.
        \item The authors should answer "Yes" if the results are accompanied by error bars, confidence intervals, or statistical significance tests, at least for the experiments that support the main claims of the paper.
        \item The factors of variability that the error bars are capturing should be clearly stated (for example, train/test split, initialization, random drawing of some parameter, or overall run with given experimental conditions).
        \item The method for calculating the error bars should be explained (closed form formula, call to a library function, bootstrap, etc.)
        \item The assumptions made should be given (e.g., Normally distributed errors).
        \item It should be clear whether the error bar is the standard deviation or the standard error of the mean.
        \item It is OK to report 1-sigma error bars, but one should state it. The authors should preferably report a 2-sigma error bar than state that they have a 96\% CI, if the hypothesis of Normality of errors is not verified.
        \item For asymmetric distributions, the authors should be careful not to show in tables or figures symmetric error bars that would yield results that are out of range (e.g. negative error rates).
        \item If error bars are reported in tables or plots, The authors should explain in the text how they were calculated and reference the corresponding figures or tables in the text.
    \end{itemize}

\item {\bf Experiments compute resources}
    \item[] Question: For each experiment, does the paper provide sufficient information on the computer resources (type of compute workers, memory, time of execution) needed to reproduce the experiments?
    \item[] Answer: \answerYes{}
    \item[] Justification: Compute settings are disclosed in the supplementary material. Training used a single NVIDIA RTX 3090 GPU for 48 hours for 1M samples. The details shown in the experiments section
    \item[] Guidelines:
    \begin{itemize}
        \item The answer NA means that the paper does not include experiments.
        \item The paper should indicate the type of compute workers CPU or GPU, internal cluster, or cloud provider, including relevant memory and storage.
        \item The paper should provide the amount of compute required for each of the individual experimental runs as well as estimate the total compute. 
        \item The paper should disclose whether the full research project required more compute than the experiments reported in the paper (e.g., preliminary or failed experiments that didn't make it into the paper). 
    \end{itemize}
    
\item {\bf Code of ethics}
    \item[] Question: Does the research conducted in the paper conform, in every respect, with the NeurIPS Code of Ethics \url{https://neurips.cc/public/EthicsGuidelines}?
    \item[] Answer: \answerYes{} 
    \item[] Justification: The research complies with the NeurIPS Code of Ethics, particularly in terms of dataset transparency and open-source release.

    \item[] Guidelines:
    \begin{itemize}
        \item The answer NA means that the authors have not reviewed the NeurIPS Code of Ethics.
        \item If the authors answer No, they should explain the special circumstances that require a deviation from the Code of Ethics.
        \item The authors should make sure to preserve anonymity (e.g., if there is a special consideration due to laws or regulations in their jurisdiction).
    \end{itemize}

\item {\bf Broader impacts}
    \item[] Question: Does the paper discuss both potential positive societal impacts and negative societal impacts of the work performed?
    \item[] Answer: \answerYes{} 
    \item[] Justification: We discuss potential applications in household robotics.
    \item[] Guidelines:
    \begin{itemize}
        \item The answer NA means that there is no societal impact of the work performed.
        \item If the authors answer NA or No, they should explain why their work has no societal impact or why the paper does not address societal impact.
        \item Examples of negative societal impacts include potential malicious or unintended uses (e.g., disinformation, generating fake profiles, surveillance), fairness considerations (e.g., deployment of technologies that could make decisions that unfairly impact specific groups), privacy considerations, and security considerations.
        \item The conference expects that many papers will be foundational research and not tied to particular applications, let alone deployments. However, if there is a direct path to any negative applications, the authors should point it out. For example, it is legitimate to point out that an improvement in the quality of generative models could be used to generate deepfakes for disinformation. On the other hand, it is not needed to point out that a generic algorithm for optimizing neural networks could enable people to train models that generate Deepfakes faster.
        \item The authors should consider possible harms that could arise when the technology is being used as intended and functioning correctly, harms that could arise when the technology is being used as intended but gives incorrect results, and harms following from (intentional or unintentional) misuse of the technology.
        \item If there are negative societal impacts, the authors could also discuss possible mitigation strategies (e.g., gated release of models, providing defenses in addition to attacks, mechanisms for monitoring misuse, mechanisms to monitor how a system learns from feedback over time, improving the efficiency and accessibility of ML).
    \end{itemize}
    
\item {\bf Safeguards}
    \item[] Question: Does the paper describe safeguards that have been put in place for responsible release of data or models that have a high risk for misuse (e.g., pretrained language models, image generators, or scraped datasets)?
    \item[] Answer: \answerNA{} 
    \item[] Justification:  Our dataset and models do not pose high misuse risks.
    \item[] Guidelines:
    \begin{itemize}
        \item The answer NA means that the paper poses no such risks.
        \item Released models that have a high risk for misuse or dual-use should be released with necessary safeguards to allow for controlled use of the model, for example by requiring that users adhere to usage guidelines or restrictions to access the model or implementing safety filters. 
        \item Datasets that have been scraped from the Internet could pose safety risks. The authors should describe how they avoided releasing unsafe images.
        \item We recognize that providing effective safeguards is challenging, and many papers do not require this, but we encourage authors to take this into account and make a best faith effort.
    \end{itemize}

\item {\bf Licenses for existing assets}
    \item[] Question: Are the creators or original owners of assets (e.g., code, data, models), used in the paper, properly credited and are the license and terms of use explicitly mentioned and properly respected?
    \item[] Answer:  \answerYes{}
    \item[] Justification:  We credit existing datasets (e.g., YOLOv8) and pretrained weights and use assets compliant with MIT and CC-BY licenses.
    \item[] Guidelines:
    \begin{itemize}
        \item The answer NA means that the paper does not use existing assets.
        \item The authors should cite the original paper that produced the code package or dataset.
        \item The authors should state which version of the asset is used and, if possible, include a URL.
        \item The name of the license (e.g., CC-BY 4.0) should be included for each asset.
        \item For scraped data from a particular source (e.g., website), the copyright and terms of service of that source should be provided.
        \item If assets are released, the license, copyright information, and terms of use in the package should be provided. For popular datasets, \url{paperswithcode.com/datasets} has curated licenses for some datasets. Their licensing guide can help determine the license of a dataset.
        \item For existing datasets that are re-packaged, both the original license and the license of the derived asset (if it has changed) should be provided.
        \item If this information is not available online, the authors are encouraged to reach out to the asset's creators.
    \end{itemize}

\item {\bf New assets}
    \item[] Question: Are new assets introduced in the paper well documented and is the documentation provided alongside the assets?
    \item[] Answer: \answerYes{} 
    \item[] Justification: We introduce a novel knolling dataset by ourselves and will release full documentation and scripts to support usage.
    \item[] Guidelines:
    \begin{itemize}
        \item The answer NA means that the paper does not release new assets.
        \item Researchers should communicate the details of the dataset/code/model as part of their submissions via structured templates. This includes details about training, license, limitations, etc. 
        \item The paper should discuss whether and how consent was obtained from people whose asset is used.
        \item At submission time, remember to anonymize your assets (if applicable). You can either create an anonymized URL or include an anonymized zip file.
    \end{itemize}

\item {\bf Crowdsourcing and research with human subjects}
    \item[] Question: For crowdsourcing experiments and research with human subjects, does the paper include the full text of instructions given to participants and screenshots, if applicable, as well as details about compensation (if any)? 
    \item[] Answer: \answerNA{} 
    \item[] Justification: This work does not involve human subjects or crowdsourcing.
    \item[] Guidelines:
    \begin{itemize}
        \item The answer NA means that the paper does not involve crowdsourcing nor research with human subjects.
        \item Including this information in the supplemental material is fine, but if the main contribution of the paper involves human subjects, then as much detail as possible should be included in the main paper. 
        \item According to the NeurIPS Code of Ethics, workers involved in data collection, curation, or other labor should be paid at least the minimum wage in the country of the data collector. 
    \end{itemize}

\item {\bf Institutional review board (IRB) approvals or equivalent for research with human subjects}
    \item[] Question: Does the paper describe potential risks incurred by study participants, whether such risks were disclosed to the subjects, and whether Institutional Review Board (IRB) approvals (or an equivalent approval/review based on the requirements of your country or institution) were obtained?
    \item[] Answer: \answerNA{} 
    \item[] Justification: his research does not involve any human subjects or studies requiring IRB approval.

    \item[] Guidelines:
    \begin{itemize}
        \item The answer NA means that the paper does not involve crowdsourcing nor research with human subjects.
        \item Depending on the country in which research is conducted, IRB approval (or equivalent) may be required for any human subjects research. If you obtained IRB approval, you should clearly state this in the paper. 
        \item We recognize that the procedures for this may vary significantly between institutions and locations, and we expect authors to adhere to the NeurIPS Code of Ethics and the guidelines for their institution. 
        \item For initial submissions, do not include any information that would break anonymity (if applicable), such as the institution conducting the review.
    \end{itemize}

\item {\bf Declaration of LLM usage}
    \item[] Question: Does the paper describe the usage of LLMs if it is an important, original, or non-standard component of the core methods in this research? Note that if the LLM is used only for writing, editing, or formatting purposes and does not impact the core methodology, scientific rigorousness, or originality of the research, declaration is not required.
    \item[] Answer: \answerNA{} 
    \item[] Justification: LLMs were not used in the methodology.
    \item[] Guidelines:
    \begin{itemize}
        \item The answer NA means that the core method development in this research does not involve LLMs as any important, original, or non-standard components.
        \item Please refer to our LLM policy (\url{https://neurips.cc/Conferences/2025/LLM}) for what should or should not be described.
    \end{itemize}

\end{enumerate}

\end{document}

%% file: sec/0_abstract.tex
\begin{abstract}
For robots to truly collaborate and assist humans, they must understand not only logic and instructions, but also the subtle emotions, aesthetics, and feelings that define our humanity. Human art and aesthetics are among the most elusive concepts—often difficult even for people to articulate—and without grasping these fundamentals, robots will be unable to help in many spheres of daily life. Consider the long-promised robotic butler: automating domestic chores demands more than motion planning; it requires an internal model of cleanliness and tidiness—a challenge largely unexplored by AI. To bridge this gap, we propose an approach that equips domestic robots to perform simple tidying tasks via knolling, the practice of arranging scattered items into neat, space-efficient layouts. Unlike the uniformity of industrial settings, household environments feature diverse objects and highly subjective notions of tidiness. Drawing inspiration from NLP, we treat knolling as a sequential prediction problem and employ a transformer-based model to forecast each object’s placement. Our method learns a generalizable concept of tidiness, generates diverse solutions adaptable to varying object sets, and incorporates human preferences for personalized arrangements. This work represents a step forward in building robots that internalize human aesthetic sense and can genuinely co-create in our living spaces.
\end{abstract}

%% file: sec/1_main.tex
\section{Introduction}
\label{sec:intro}

Designing a robot for household tasks has always presented unique challenges \cite{zachiotis2018survey, kim2019control,zhong2021gap,batra2020rearrangement}. Unlike industrial settings characterized by uniformity and limited object variety, household environments are filled with diverse objects. Recent progress in object rearrangement has shown that robots can organize objects in partially arranged scenes, cluttered tabletops, and constrained environments\cite{ramachandruni2023consor,gao2023effectively,lou2023adversarial}. While their work has provided effective approaches to solving task-specific problems, it is crucial to develop methods that can learn a generalized representation of tidiness. Our motivation is to enable robots to understand and apply tidiness concepts that improve with more data, unlike traditional rule-based methods that become increasingly complex with more scenarios.

In common daily environments, including both local environments like a desk, and larger environments like a house, the objects encountered vary constantly. In such dynamic settings, providing specific goals for each object is impractical and limits the robot's generalizability. Instead, robots should be capable of organizing the environment without relying on specific target positions or constant human supervision. This requires the robot to understand and apply a broad concept of organization that extends beyond the specifics of any single task. The necessary adaptability is similar to human-like cognition, where decisions are made not only based on spatial arrangement but also considering the aesthetic and functional aspects of tidiness.

\begin{figure}[t]
    \centering
    \includegraphics[width=0.9\textwidth]{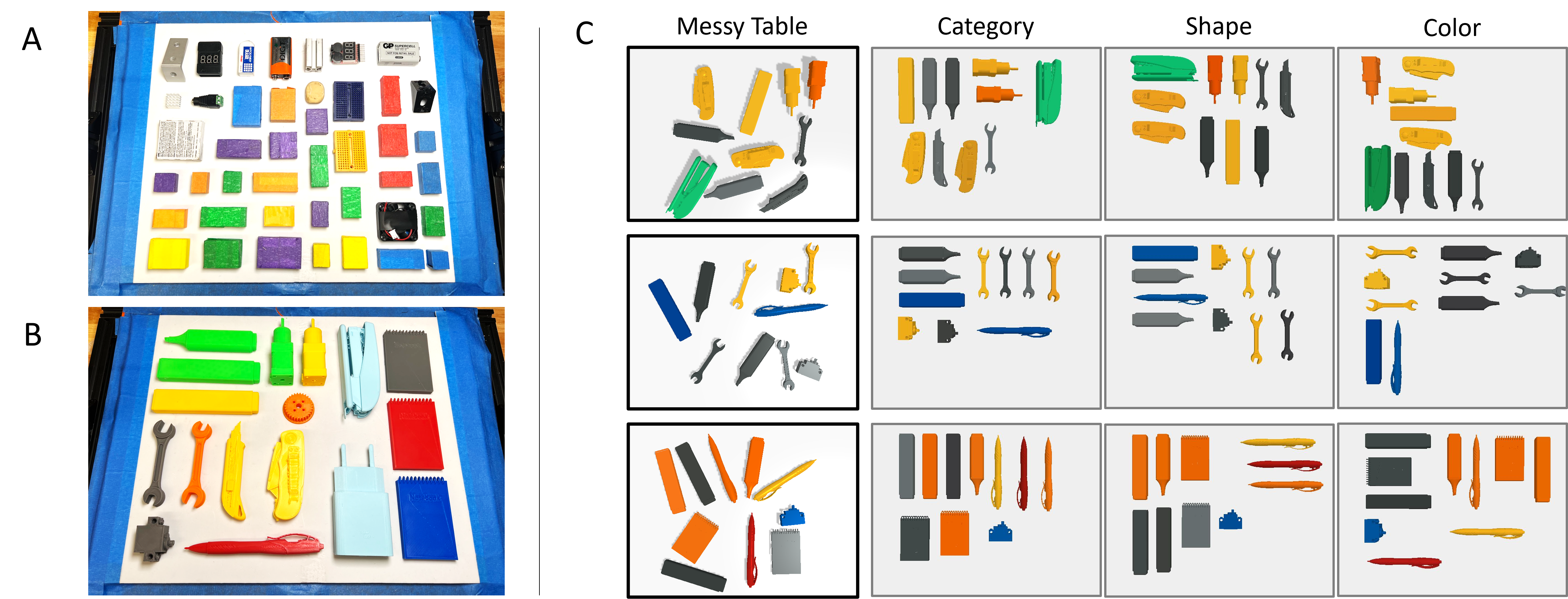} 
    \caption{\textbf{Examples from the knolling task.} 
    A) A batch of small items, including daily necessities. B) A batch of big items fabricated by a 3D printer. 
    C) Diverse knolling preferences demonstrated in experiments. The model adapts its tidying behavior based on different preferences: object category (group by function), size (large objects first), and color (group by hue).}
    \vspace{-10pt}
    \label{fig: obj}
\end{figure}

``Knolling'' is one such concept intrinsic to humans, referring to the intuitive ability to organize items in a manner that is both aesthetically pleasing and space-efficient (Fig. \ref{fig: obj}). Our solution lies in decoupling the cognitive model, which encapsulates the representation of knolling, from the other modules, such as the visual perception system and motor controllers\cite{jang2017end,mousavian20196}. This division enhances the modularity and interpretability of the entire system compared to only training a single policy mapping from observation to action.

\begin{figure*}[t]
    \centering
    \includegraphics[width=0.99\textwidth]{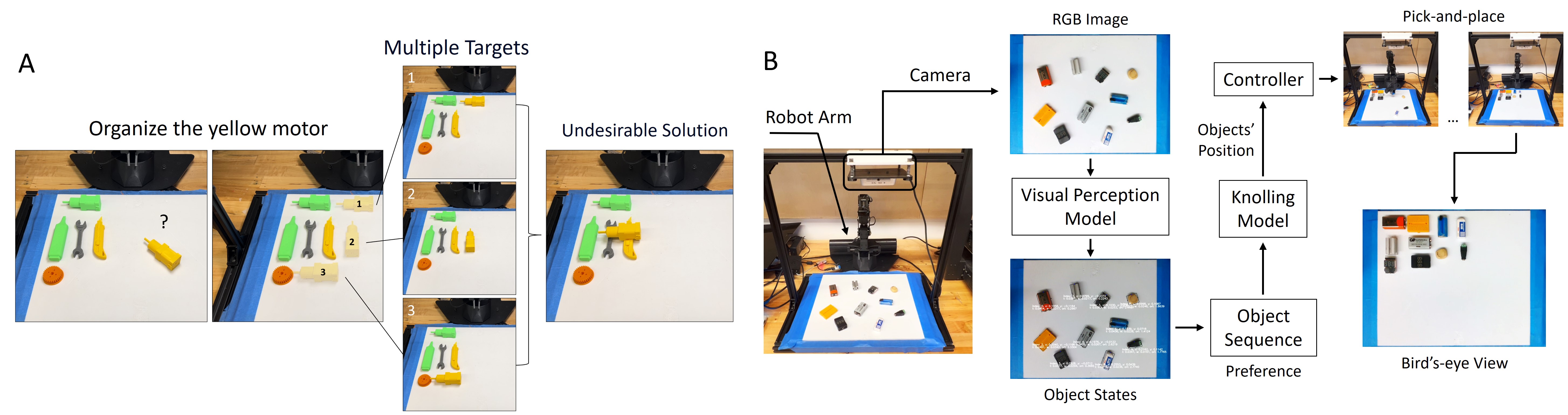}
    \caption{\textbf{A) Challenges in rearrangement tasks with multiple solutions.} From left to right: the initial state of the work area with an unplaced yellow motor. Three proposed placement options (1, 2, 3) for the yellow motor, considering factors such as object category, color similarity, and spatial efficiency. Placement by a regression-based model optimizing for minimal loss across three solutions, leading to undesirable results: placing the motor on a utility knife. \textbf{B) Knolling pipeline.}  The left side of the figure displays a cluttered desktop with various objects such as batteries, erasers, electronic components, and other daily necessities in the lab. Our robot initiates the knolling process after detecting and identifying these objects through a camera. The right side of the figure depicts the outcome of this process, presenting a tidy, well-organized desktop. This transformation exemplifies the robot's ability to apply the knolling model, execute a tidying task, and create a pleasing and space-efficient arrangement.}
    \vspace{-10pt}
    \label{fig:pipeline}
\end{figure*}

Given the subjectivity of tidiness and the multiple optimal solutions that could satisfy different individuals' preferences, knolling is an abstract problem without any definitive standards. Consider the scenario depicted in Fig.\ref{fig:pipeline}A, where a robotic arm performs a task to integrate a yellow motor into a partially organized environment. The decision on where to place the motor can be depended on various factors, including object category, color, and the need for space efficiency. Training a model solely on regression to predict target positions based on object states might lead to local optimal outcomes. For instance, in Fig.\ref{fig:pipeline}A a model optimized to minimize the loss across different placement preferences could converge to an average of the three potential target positions, resulting in the motor being placed in a position that is overlapped with the utility knife. This is just an example of a one-step rearrangement of a partially organized scenario. A wrong-placed object may cause the next step of rearrangement to become worse because the errors have accumulated. Moreover, quantifying tidiness with a singular metric is challenging, as no simple equation can encapsulate the diverse aspects of what constitutes a ``tidy" space.

In this work, we propose a self-supervised learning framework for modeling tidiness through demonstrations, akin to training a Large Language Model (LLM) with human conversational datasets\cite{chen2021smile,zeng2020tossingbot,achiam2023gpt,radford2019language}. We draw parallels between knolling objects and language, where objects act as individual ``words" that can be combined in various ways to form ``sentences'' with identical meanings. Besides, the transformer architecture can handle varying input and output sizes with autoregression, which fits the knolling task as the number of objects in the environment is arbitrary\cite{vaswani2017attention,gillioz2020overview}. Leveraging the Gaussian Mixture Model (GMM), we address the multi-label prediction challenges inherent in the knolling task, where multiple valid placements may exist for the same object depending on contextual preferences (e.g., spatial efficiency or category)\cite{reynolds2009gaussian}. 

Our research contributes by equipping robots with a conceptual understanding of tidiness, demonstrating an integrated pipeline for knolling in real-world tasks. As shown in Fig.\ref{fig:pipeline}B , itcomprises three stages: a knolling model, a visual perception model, and a robot arm controller. The knolling model, based on a transformer architecture, predicts the target positions of objects. The visual perception model, based on our customized YOLO v8, detects objects from an RGB image \cite{reis2023real}. Finally, the robot arm controller, leveraging the outputs from the previous stages, guides the robot arm to execute the knolling tasks.

The key contributions of this work are as follows:
\begin{enumerate}[leftmargin=*]
  \item We propose a novel self-supervised learning framework for modeling the representation of tidiness. This allows the model to learn generalizable tidiness patterns directly from demonstrations, improving its performance as more data becomes available.
  \item We leverage transformer architectures and Gaussian Mixture Models to address the multi-label prediction challenges inherent in knolling tasks, and can process unseen objects and preferences, highlighting the potential for deploying such systems in dynamic household environments.
  \item We demonstrate an integrated pipeline combining the learned knolling model with visual perception and robotic control modules. This allows for deploying knolling capabilities on a real robotic system to organize cluttered environments with varying object quantities and types.
  \item We contribute to the research community by not only open-sourcing our dataset of tidy object arrangements but also establishing a benchmark with comprehensive evaluation metrics for the knolling task. This benchmark will enable comparative studies and further exploration of arrangement tasks with arbitrary object numbers and shapes.
\end{enumerate}

\section{RELATED WORK}
The domain of robot learning and manipulation tasks has seen significant advancements in recent years. Various manipulation tasks, such as wiping and polishing, stacking, peg-in-hole, and pick-and-place tasks, have been explored\cite{berscheid2019robot,zhu2020robosuite,leidner2019cognition,8593647}. In the realm of stacking tasks, Lee et al. used offline reinforcement learning (RL) to improve upon existing policies for robotic stacking of objects with complex geometry, while Furrer et al. proposed an algorithm for suggesting stable poses for stacking, validated through a real-world autonomous stacking workflow\cite{lee2021beyond,7989272}. For pick-and-place tasks, Gualtieri et al. proposed a deep reinforcement learning approach, and Zeng et al. introduced the Transporter Network for vision-based manipulation tasks\cite{schoettler2020meta, zeng2021transporter}. 

In the domain of machine learning models, transformers have revolutionized many fields with their attention-based architecture\cite{radford2018improving,khan2022transformers,liu2022structformer,kapelyukh2023dall,wei2023lego,liu2022structdiffusion}. The success of transformer models in NLP has prompted exploration in their application to robotic tasks\cite{kim2021transformer,dasari2021transformers,ren2024neural}. Jangir et al. proposed the use of transformers with a cross-view attention mechanism for effective fusion of visual information from two cameras for RL policies\cite{9690036}. Zhu et al. introduced VIOLA, which used a transformer-based policy to improve the robustness of deep imitation learning algorithms\cite{zhu2023viola}. Shridhar et al. proposed PerAct, which used the preceiver transformer to encode language goals and RGB-D voxel observations\cite{shridhar2023perceiver}. Jain et al. proposed the Transformer Task Planner, which could be pre-trained on multiple preferences, and Dasari et al. explored one-shot visual imitation learning using the transformer architecture\cite{jain2023transformers}.

In the object rearrangement field, most recent work has focused on rearranging objects based on explicit instructions\cite{goodwin2022semantically,cheong2020relocate}. Some other works use a more general method to train a representation model or reward function to supervise the object rearrangement policy without explicit human supervision\cite{wu2022targf,wang2022generalizable}. With the rapid rise of LLM as powerful all-purpose models, many works have leveraged them to processs\cite{liang2023code,blukis2022persistent,ahn2022can}. Instead of giving natural language instructions to the robot, some other works train a robot a common sense of how to place the objects through demonstrations\cite{jain2023transformers,kapelyukh2022my,sarch2022tidee}. Combining both approaches, Housekeep\cite{kant2022housekeep} provides a method that leverages LLM to train a common sense of tidiness for robots.
While these works have contributed significantly to robot manipulation, we aim to provide a common sense of tidiness for robots through demonstrations. We focus on knolling tasks that rearrange messy objects on a desktop into a neat layout in the real world. Our approach allows for the handling of varying object types, sizes, and quantities, thus enabling the robot to generate an aesthetically pleasing and space-efficient arrangement of items, similar to human performance. 

\begin{figure*}[h]
    \centering
    \includegraphics[width=0.9\textwidth]{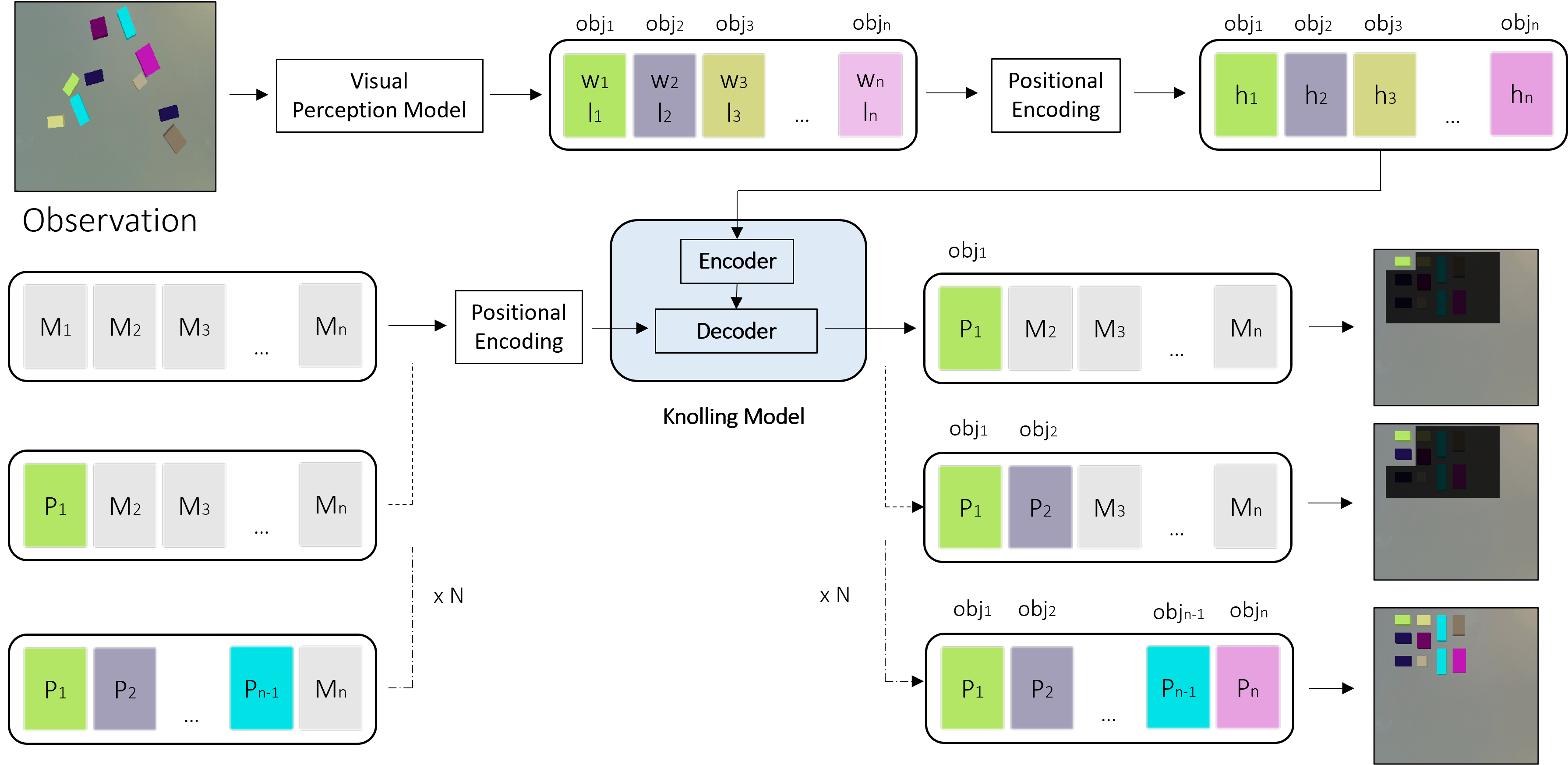}
    \caption{Knolling Model Learning Framework: The pipeline begins with the visual perception model, which processes an input image to identify objects and extract their state representations—including width (w), length (l), position and orientation, presented as a list. However, only w and l are used as input for the knolling. The knolling model takes high-dimensional object states (h) derived from positional encoding as input. During training, a masked learning approach is employed, where part of the object data (M) is masked, and the model learns to predict the next object's position (P). The model predicts N target positions after N iterative processes.}
    \vspace{-10pt}
    \label{fig:learning_f}
\end{figure*}

\section{METHOD}

\subsection{Data Representation and Generation}
We represent objects solely based on their dimensional parameters (width and length), as these are the fundamental properties governing spatial organization. semantic attributes like color and category are excluded from the model input due to the subjective variability and potential bias they introduce in the context of tidiness. Unlike quantifiable features (e.g., position, size), attributes like color and category are often interpreted differently based on context and individual perception. Focusing on spatial and geometric attributes is more objectively measurable. This approach helps the model generalize tidiness based on universally applicable metrics rather than subjective semantic interpretations that might vary across different settings.
To generate a diverse dataset of tidy arrangements, we design an optimization strategy that iteratively adjusts object placements to minimize the occupied area on the table. This strategy can control the results by adjusting placements based on object attributes such as color, dimension, or category, yielding distinct organizational patterns and aesthetic preferences. By iterating through multiple configurations, this stochastic process produces 2.4 million demonstrations that span a broad spectrum of tidiness concepts and visual styles, providing a rich dataset for training.
For objects with irregular shapes (Fig. \ref{fig: obj}B), we train a visual perception system to perform segmentation and compute the minimum bounding box, enabling our framework to accommodate unconventional geometries.

\subsection{Training The Knolling Model}
Our self-supervised learning framework for training knolling models follows a curriculum-based approach, progressing from simple to complex tasks as shown in Fig. \ref{fig:learning_f}. We designed a two-stage training process to first acquaint the model with the basics of object arrangement and then refine its proficiency in executing complete knolling tasks from scratch. By training a representation of tidiness through self-supervised learning, our model's performance improves as more data becomes available, unlike traditional rule-based methods that become increasingly complex with more scenarios. For further details on the two training phases, pre-training with self-supervised Learning, amd finetuning, please see Appendix L.
Our knolling pipeline comprises three modules: a knolling model, a visual perception model, and the robot arm controller for executing the pick-and-place task. For further details on the pipeline and training objectives, see Appendix K.

\section{Experiments}
\label{sec:result}
To evaluate the performance and practical applicability of our proposed knolling framework, we conducted experiments in both simulated and real-world environments. Additionally, we performed quantitative analyses and ablation studies to validate the effectiveness of our approach and elucidate the contributions of different components.

\begin{figure*}[h]
\begin{center}
    \includegraphics[width=0.99\textwidth]{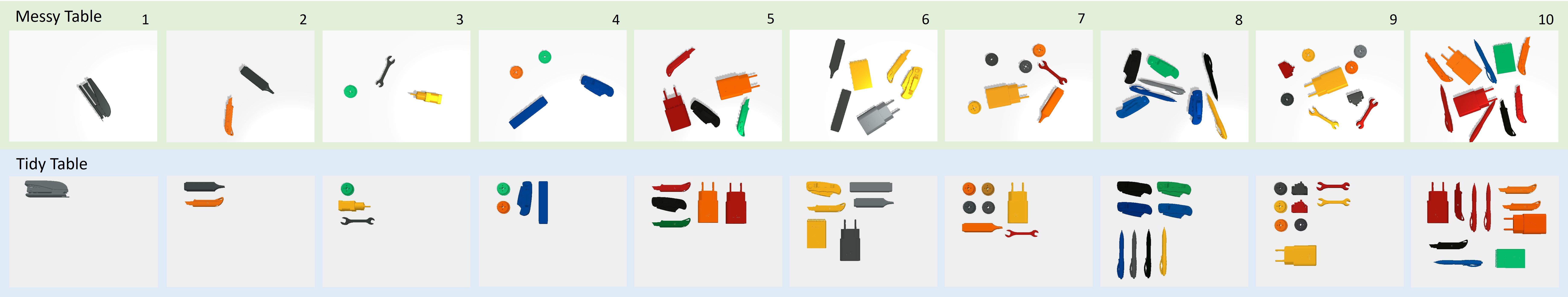}
\end{center}
    \caption{Examples of knolling messy tables with different numbers of objects. The figure shows ten examples of tables before and after the knolling process in the simulation.
    }
    \label{fig:MvsT}
\vspace{-10pt}
\end{figure*}

\subsection{Qualitative Evaluations in the Simulation}
\textbf{Handling Varying Object Quantities.}
In real-world scenarios, the number of objects on a surface can vary significantly. To assess our model's ability to handle this variability, we conducted experiments with diverse object counts ranging from 2 to 10 items. Our model, leveraging its autoregressive nature, successfully generated tidy arrangements across all object quantities, as shown in Fig. \ref{fig:MvsT}. This adaptability to varying input sizes is a crucial aspect of our approach, enabling it to generalize to different scenarios without explicit retraining.

\textbf{Generating Diverse Solutions Based on Preferences.}
Tidiness preferences can differ significantly among individuals, leading to multiple valid solutions for the same set of objects. To capture this diversity, our model decouples preferences from the input data (which consists solely of object dimensions) and instead encodes preferences through the order of the input sequence. By altering the input order based on criteria such as object category, color, or size, our model generates distinct tidy arrangements tailored to specific preferences, as illustrated in Fig. \ref{fig: obj} C. This flexibility to incorporate user preferences without modifying the model architecture is a key strength of our approach.


\begin{figure*}[t]
    \centering
    \includegraphics[width=0.99\textwidth]{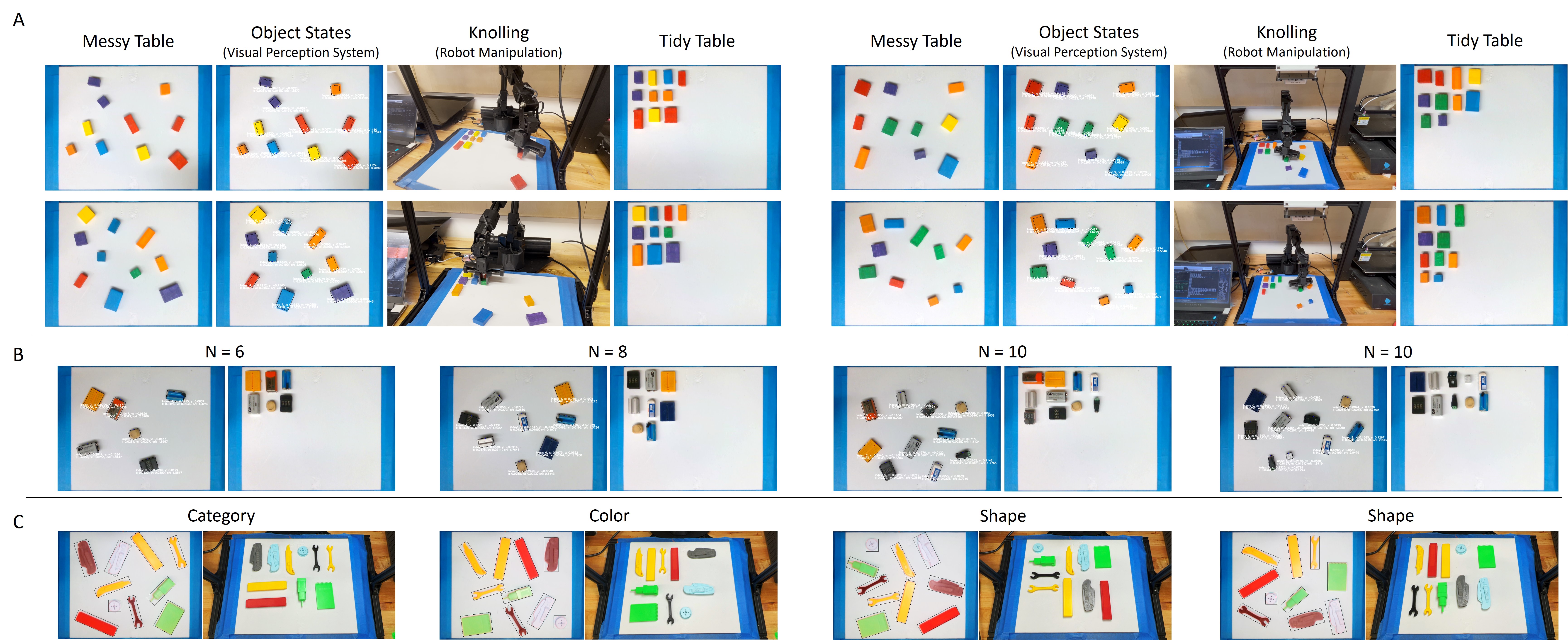}
    \caption{ \textbf{A)} Box knolling in the real world. In each test, we show four columns. Column 1: The initial state of the objects on a table, as captured by the overhead camera. Column 2: The same scenario as Column 1, with added key points and contour outlines indicating the detected objects. Column 3: Action snapshot of the robot executing the knolling task. Column 4: The final state of the workspace post-knolling, presenting an tidy table. \textbf{B)} Real-world Knolling Process with Different Object Numbers. This figure exhibits the practical application of our knolling model in four diverse scenarios. Each column corresponds to a different setup with a distinct number of objects (6, 8, 10, and 10). We show the initial messy state captured by the overhead camera and the organized layout after the knolling task is completed by the robot arm. These comparative visuals underline our robot's proficiency in performing real-world knolling tasks across varied object quantities. \textbf{C)} For the same objects on the table, our robot can perform knolling tasks with different solutions based on preferences based on category, color, or shape.}
    \vspace{-10pt}
    \label{fig:real_test}
\end{figure*}

\subsection{Real-World Knolling Experiments}
To demonstrate the practical applicability of our framework, we deployed our trained knolling model on a 5-DoF robotic arm (WidowX 200) equipped with an Intel Realsense D435 camera for visual perception. We randomly placed 6-10 boxes of varying sizes and colors within the robot's workspace, simulating cluttered tabletop scenarios.
The integrated pipeline (Fig. \ref{fig:pipeline}) first captures an overhead image of the cluttered scene, which is processed by our visual perception module to detect objects, perform segmentation, and extract their dimensions and poses. The knolling model then predicts the target positions for each object based on their dimensions. Finally, the robotic arm controller executes smooth pick-and-place operations, guided by the predicted target positions and current object poses, to realize the tidy arrangements.
Fig. \ref{fig:real_test}A showcases real-world knolling tasks performed on different box configurations, demonstrating our system's capability to organize diverse object sets. Fig. \ref{fig:real_test}B further illustrates the system's adaptability to varying object quantities, successfully knolling scenes with 6, 8, and 10 objects. Additionally, Fig. \ref{fig:real_test}C highlights the generation of distinct tidy arrangements for the same set of objects, reflecting different user preferences based on category, color, or shape. We encourage readers to view our supplementary video for a comprehensive visualization of these real-world knolling experiments. For further quantitative evaluation and an ablation study, please see Appendix I. For an analysis of how performance scales with data, see Appendix J.

\section{CONCLUSIONS}
In this work, we introduced a self-supervised learning framework for enabling robots to understand and replicate the representation of tidiness, or ``knolling," from demonstrations of well-organized object arrangements.
Our approach leverages transformer architectures and Mixture Density Networks to model the inherent multi-modality of knolling tasks, where identical object configurations can lead to diverse but tidy arrangements. Our model learns a generalizable representation of tidiness that can adapt to varying object quantities and incorporate user preferences by training on a large dataset of knolling demonstrations without explicit human supervision or target positions.
We demonstrated the effectiveness of our framework in generating aesthetically pleasing and space-efficient object arrangements. Our quantitative evaluations showed that our transformer-based model consistently outperformed baseline architectures, highlighting the advantages of self-attention and autoregressive mechanisms for handling variable input sizes and multi-label predictions.
Moreover, we presented an integrated pipeline that combines our learned knolling model with visual perception and robotic control modules, enabling the deployment of knolling capabilities on a real robotic system. Our real-world experiments showcased the system's ability to organize cluttered environments with diverse object types and quantities, adapting to different user preferences based on criteria such as object category, color, or shape.
This work represents an important step forward in imbuing robots with human aesthetics, a critical requirement for effective human-robot interaction, and we hope our work inspires further research in this area.


%% file: sec/X_suppl.tex
\clearpage

\section*{Technical Appendices}
\section{Training Objectives}
We employ a combination of loss functions during training:
\subsubsection*{1. Log-likelihood Loss ($L_{\text{ll}}$)}
The Log-likelihood loss is a standard approach when working with GMM, which allows the model to express uncertainty and multi-targets in the predictions by assigning probabilities to a range of possible positions for each object. This is particularly useful for handling the diversity and complexity of knolling tasks, where the spatial arrangement of objects can vary significantly. 
The log-likelihood loss for each object $i$, considering a Gaussian Mixture Model (GMM) with $J$ components, is given by:
\begin{equation}
L_{\text{ll}} = -\sum_{i=1}^{N} \log \left( \sum_{j=1}^{J} p_{ij} \cdot \mathcal{N}(S_i | \mu_{ij}, \sigma_{ij}^2) \right)
\end{equation}
where $N$ is the total number of objects, $p_{ij}$ is the weight of the $j$-th component for the $i$-th object, $\mathcal{N}(S_i | \mu_{ij}, \sigma_{ij}^2)$ is the probability density of the $i$-th object's target position $S_i$ under the $j$-th Gaussian component with mean $\mu_{ij}$ and variance $\sigma_{ij}^2$.

\subsubsection*{2. MSE Loss ($L_{\text{MSE}}$)}
The Mean Squared Error (MSE) Loss computes the squared difference between the sampled predicted position ($\hat{S}$) and the target position ($S$):
\begin{equation}
L_{\text{MSE}} = \frac{1}{N} \sum_{i=1}^{N} (\hat{S}_i - S_i)^2
\end{equation}
where $\hat{S}_i$ is the sampled position from the predicted distribution for the $i$-th object, and $S_i$ is the actual target position. 

\subsubsection*{3. MSE Min Loss ($L_{\text{MSE Min}}$)}
This encourages the model not only to predict accurately but also to account for the multiple solutions in the dataset.
Considering \(J\) components in the Gaussian mixtures for each prediction step, the MSE Min Loss is calculated as follows:
\begin{equation}
L_{\text{MSE Min}} = \frac{1}{N} \sum_{i=1}^{N} \min_{j=1}^{J} \left( (\hat{S}_{ij} - S_i)^2 \right)
\end{equation}
In a multi-target task, relying solely on MSE could lead the model to predict average positions in the presence of ambiguity, which is not desirable. The MSE Min Loss addresses this by allowing the model to choose the prediction (from among multiple Gaussian components) that is closest to the actual target position.

\subsubsection*{4. Overlapping Loss ($L_{\text{overlap}}$)}
The Overlapping Loss penalizes predictions where objects overlap or cross boundaries. It acts as a form of regularization but with a very specific purpose: to ensure that the predicted positions do not result in physically impossible arrangements, such as overlapping objects or objects placed outside the designated workspace. This loss component is essential for maintaining the feasibility of the predicted knolling arrangements, ensuring that the model's predictions adhere to spatial constraints inherent in the real-world application of knolling. The loss function can be expressed as:
\begin{equation}
L_{\text{overlap}} = \sum_{i \neq k} \text{Penalty}(\hat{S}_i, \hat{S}_k, w_n, l_n, W, L)
\end{equation}
The penalty function calculates the overlap and out-of-boundaries areas based on their predicted positions ($\hat{S}_i, \hat{S}_j$), the given dimensions of objects ($w_n, l_n$) and workspace ($W, L$).

\subsubsection*{5. Entropy Loss ($L_{\text{entropy}}$)}
This loss function encourages diversity in the predictions by measuring the entropy across the probabilities of the J Gaussian components for each prediction. The entropy loss is defined as:

\begin{equation}
L_{\text{entropy}} = -\sum_{i=1}^{N} \sum_{j=1}^{J} p_{ij} \log(p_{ij})
\end{equation}

\subsubsection*{6. Weights of Loss Components}
Each loss component was assigned a specific weight to balance their contributions during training, with the aim of optimizing knolling performance. The weights were determined empirically, through iterative experimentation and performance evaluation, to ensure the model adequately minimized both positional errors and object overlap.

\section{Iterative Optimization for Knolling Dataset Generation}
While the rule-based arrangement method presented here provides a systematic approach to generating a knolling dataset, it also has inherent limitations. Rule-based methods excel at articulating specific, human-defined rules of neatness—such as alignment, spacing, and categorization by color or type—that are relatively straightforward to parameterize. However, much of neatness is inherently subjective and nuanced. Although humans can often recognize whether an arrangement is “neat” or “not neat,” they may struggle to articulate the underlying principles that define these notions. This subjectivity poses a challenge for rule-based systems, which can only operate within explicitly defined rules and fail to capture the subtleties of more subjective or abstract concepts of neatness.

Defining an exhaustive set of rules to account for all possible tidiness concepts is infeasible, as the number of potential layouts grows exponentially with object variety, user preferences, and spatial constraints. As a result, rule-based or optimization methods alone cannot cover all valid solutions. Instead, we take the opposite approach, learning a generalizable representation of tidiness from demonstrations through self-supervised learning.

While rule-based optimization is valuable for data generation, it is often computationally expensive and inefficient for real-time object arrangement. In contrast, the learned knolling model operates in a single forward pass, making predictions significantly faster than iterative optimization. This efficiency enhances scalability, enabling practical deployment in real-world robotic applications.

The rule-based optimization bootstraps the learning process, providing structured training examples to establish an initial understanding of tidiness. However, it is not a substitute for data-driven method. The model leverages this structured data as a foundation but learns beyond it, capturing nuanced spatial relationships and placement strategies that a purely optimization-based approach could never explicitly define.

\subsection{Objects Used in the Knolling Task}
we provide the following details regarding our dataset.
\textbf{Rectangular Objects:} A subset of the dataset consists of randomly generated rectangular objects, with their length and width sampled from a predefined range. These synthetic objects serve as fundamental elements in training the model to generalize tidiness principles. Some of these generated objects were physically printed and used in real-world experiments to validate the learned representations. \textbf{Laboratory Items:} In addition to the rectangular objects, we incorporated commonly used laboratory items such as batteries, erasers, and electronic components. These objects were selected to introduce variation in shape and appearance. To facilitate color preference learning, we 3D-printed different versions of these objects in multiple colors. This ensures that the model can account for color-based grouping in tidiness preferences.

\subsection{Arrangement Strategy and Policy Setup}
To generate a knolling dataset, a meticulous arrangement framework is developed, which integrates parameterized object properties, multiple preferences, and an iterative layout optimization process. This structured dataset generation ensures that the knolling model captures diverse object configurations, achieving precise arrangements. The framework enables training on a fixed number of objects (e.g., 10 objects), while flexibility in handling arbitrary numbers during deployment is achieved through masking techniques.

The knolling framework begins by setting arrangement policies that define the characteristics and layout behavior of each object. These policies include:

\textbf{Object Dimensions:} Length, width, and height ranges for each object are specified. For example, the length range is set to [0.036, 0.06] m, the width range to [0.016, 0.036] m, and the height range to [0.01, 0.02] m. This range allows for a controlled variation in object sizes within a standard workspace.

\textbf{Classification Criteria:} Objects are grouped based on attributes like type and color. Each classification criterion adds to the combinatorial complexity, as multiple preferences, creating distinct layouts can group objects.

\textbf{Preferences and Policies:} Policies for even distribution, forced alignment, and iteration constraints are set to control spatial relationships. These preferences guide the arrangement structure, affecting how objects are spaced, oriented, and aligned in the final layout.

This policy setup defines a complex framework where each additional preference increases layout complexity. Expanding beyond ten objects or adding more preferences exponentially intensifies the dataset generation's computational requirements.

\subsection{Arrangement Iteration}
Object data, such as length, width, height, class, and color, are generated randomly, guided by the predefined limits set in the arrangement policies. With initial object attributes and groupings established, an iterative process is applied to optimize object layout. Iterations focus on minimizing the space occupied while meeting arrangement constraints, such as spacing, alignment, and rotational preferences. For each iteration, factors of the object count are calculated to determine potential grid configurations. For example, an object count of 10 allows configurations like a 2x5 grid, which are systematically tested. Objects are arranged iteratively in each grid configuration. During each iteration, the layout is scored based on spatial minimization and adherence to the arrangement preferences. Across iterations, the layout with the minimal space occupation and best policy adherence is selected as the final arrangement. Iteration count determines the thoroughness of this search, with higher values leading to more refined arrangements. 

To improve computational efficiency, a threshold-based early-stopping mechanism is integrated. A threshold score for spatial occupation is predefined within the policy. If a layout configuration meets or surpasses this threshold, the iteration process halts early, reducing the total number of iterations. This early stopping criterion is crucial in managing computational resources, especially as the number of objects and arrangement preferences increases.

\section{Comparison of Knolling Bot and Related Methods}
\begin{figure*}[h]
    \centering
    \includegraphics[width=0.99\linewidth]{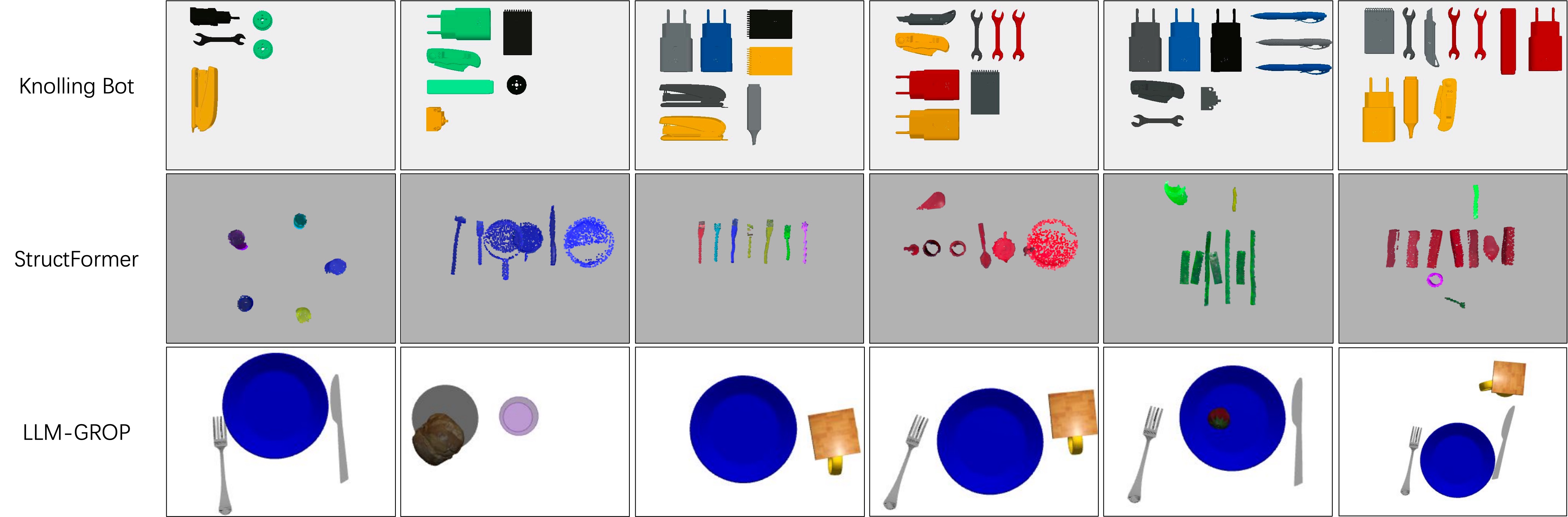}
    \vspace{-5pt}
    \caption{Comparison between our method and other methods. Our model can rearrange an arbitrary number of objects based on color, size, or categories and provide neat layouts.}
    \label{fig:comparison}
    \vspace{-15pt}
\end{figure*}

In this section, we compare our Knolling Bot approach with other notable methods in robotic object rearrangement, specifically StructFormer and LLM-GROP. A visual comparison is provided in Fig. \ref{fig:comparison}. Our method learns generalizable tidiness patterns through a self-supervised learning framework, enabling it to rearrange an arbitrary number of objects into neat and visually organized layouts. This approach allows the robot to autonomously organize objects based on attributes such as color, size, or category, resulting in aesthetically pleasing and space-efficient arrangements. StructFormer employs a transformer-based neural network that interprets structured language commands and partial-view point clouds to rearrange objects into specified configurations. While effective in executing instructions like forming circles or lines, its reliance on explicit commands may limit its adaptability to more complex or unstructured tidiness preferences. Additionally, StructFormer's focus on predefined structures may not fully capture the nuances of human-like organization, and its scalability to larger or more diverse object sets has not been extensively demonstrated. LLM-GROP leverages large language models (LLMs) combined with task and motion planning to achieve semantically valid object rearrangements in tabletop scenarios. It utilizes LLMs to extract commonsense knowledge about object configurations and integrates this with a task-motion planner to adapt to varying scene geometries. However, as demonstrated in its experiments, LLM-GROP is designed for tasks involving a limited number of objects, typically three to five, and focuses on specific tasks such as setting a dinner table. In contrast, Knolling Bot's self-supervised learning framework enables it to handle a larger and more variable number of objects without explicit instructions, making it more versatile in organizing diverse and unstructured environments. Notably, Knolling Bot achieves a level of neatness and organization that closely aligns with human aesthetic preferences, arranging objects in a manner that is both visually pleasing and functionally efficient. This human-like understanding of tidiness distinguishes Knolling Bot from other methods

\vspace{-8pt}

\section{Gaussian Mixture Model for Multi-target Learning}

To address the multi-target nature of knolling tasks, where multiple valid arrangements are possible for a given set of objects as shown in Fig.\ref{fig:pipeline}A, we employ Mixture Density Networks (MDNs) to model the potential distribution of object placements. MDNs combine a neural network with a Gaussian Mixture Model (GMM), allowing the network to predict not just a single target position, but a weighted mixture of Gaussian distributions representing the diverse set of plausible positions for each object.

Formally, the MDN component of our model predicts the parameters ${(\pi_j, \mu_j, \sigma_j)}_{j=1}^J$ for a GMM with $J$ components, where $\pi_j$ are the mixing coefficients, $\mu_j$ are the means, and $\sigma_j$ are the standard deviations of the $J$ Gaussian distributions. For each object $i$, the predicted probability density function (PDF) over its possible placement positions $S_i$ is given by:

\begin{equation}
p(S_i) = \sum_{j=1}^J \pi_{ij} \mathcal{N}(S_i | \mu_{ij}, \sigma_{ij}^2)
\end{equation}

This formulation allows the model to capture the multiple knolling solutions, where identical object configurations can lead to different but equally valid tidy arrangements. During inference, we can sample from the predicted GMM to obtain diverse knolling predictions, or select the component with the highest mixing coefficient for a single deterministic prediction. Incorporating MDNs into our transformer architecture prevents the model from converging to averaging solutions and captures the inherent uncertainty and variability in tidiness preferences.

\section{Position Encodering}
Our model utilizes two position encoding methods to process the input data effectively. The first method, aligned with the original transformer paper's approach, assigns unique tokens to each position in the list. This gives the data a sense of order, enabling the model to predict the target position of the initial objects and apply autoregression for the remaining objects. The second position encoding method maps the input data into a higher-dimensional space via sinusoidal functions. This method allows the model to handle higher frequency representations that potentially improve performance. The input, originally a 2-dimensional vector representing object length and width, is transformed into a 21-dimensional feature vector through this encoding method. By varying the wavelength at five different frequencies, the position encoding captures fine-grained patterns in the data, leading to more accurate predictions.

\section{Transformer Performance across Dataset Sizes}

This experiment aimed to evaluate the performance of our knolling model across different scales of data. Four dataset sizes were selected for this study: 125k, 250k, 500k, and 1M. Each variant of the knolling model was trained independently until convergence. Following training, all models were evaluated using a consistent test dataset to maintain uniformity in the assessment.

The test errors obtained for each dataset size are shown in Table \ref{tab:sup_experiment}. The results highlight a consistent decline in error as the dataset size augments. This trend validates the widely accepted belief that transformers excel with the increase in data volume.

\section{Visual Perception Model Evaluation}

\subsection{YOLO Segmentation Model}
Our visual perception module employs a customized YOLO v8 model, fine-tuned for object detection, segmentation, and attribute prediction. We trained and evaluated two versions of the model: one for the simulated environment (YOLO-sim), and another for real-world deployment, (YOLO-real).

\subsubsection{Training Settings}
\textbf{Simulation Model (YOLO-sim):}
\begin{itemize}
\item Number of categories: 9
\item Number of objects per image: 4-11
\item Dataset sizes: 3200 train images, 800 test images
\item Pre-trained model: yolov8n-seg
\end{itemize}

\textbf{Real-World Model (YOLO-real):}
\begin{itemize}
\item Number of categories: 9
\item Number of objects per image: 4-11
\item Dataset sizes: 80 train images, 20 test images
\item Pre-trained model: YOLO-sim
\end{itemize}

\subsubsection{Evaluation Metrics and Results}
We evaluated the models' performance using the following metrics:
\begin{itemize}
\item Average Accuracy: Measures the accuracy of category predictions.
\item Confusion Matrix: Provides a detailed breakdown of the model's category prediction performance.
\item Average Hamming Distance: Assesses the quality of the predicted segmentation masks compared to the ground truth.
\end{itemize}

The evaluation results for both models are summarized in Tab.\ref{tab:yolo_eval}.

\begin{table}[h]
\centering
\caption{YOLO Segmentation Model Evaluation Results}
\label{tab:yolo_eval}
\begin{tabular}{|l|c|c|}
\hline
\textbf{Metric} & \textbf{YOLO-sim} & \textbf{YOLO-real} \\
\hline
Average Hamming Distance & 0.06743 & 0.05831 \\
Average Accuracy & 0.999 & 0.9857 \\
\hline
\end{tabular}
\end{table}

\begin{figure}[t]
    \centering
    \includegraphics[width=0.98\textwidth]{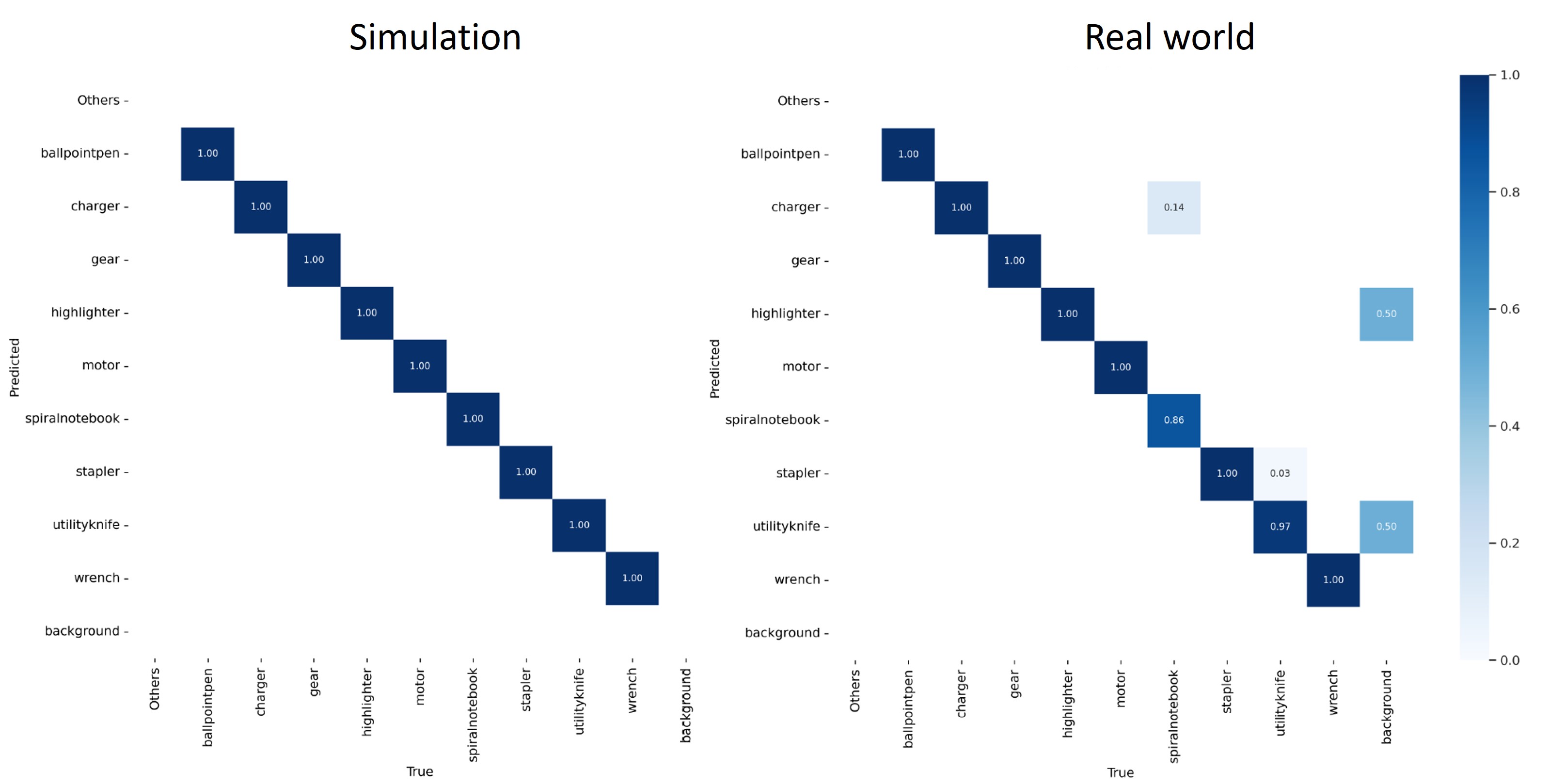}
    \caption{Confusion matrix for the YOLO segmentation model. The matrix shows the model's category prediction performance, with each cell representing the number of instances predicted as a particular category (column) compared to the true category (row). The diagonal cells indicate correct predictions, while off-diagonal cells represent misclassifications.}
    \vspace{-10pt}
    \label{fig:confusion}
\end{figure}

The results demonstrate that both models achieve high accuracy in category prediction and generate precise segmentation masks, with the real-world model slightly outperforming the simulation model in terms of Average Hamming Distance. The confusion matrices for both models are provided in Fig.\ref{fig:confusion}.

These evaluation results validate the effectiveness of our visual perception module in accurately detecting, segmenting, and categorizing objects in both simulated and real-world environments, enabling robust input for the subsequent knolling pipeline.

\section{Comparison to Baselines}
\label{sec:Baselines}
To further validate the effectiveness of our transformer-based knolling model, we conducted additional experiments, exploring the model’s performance relative to widely used architectures, including MLP, LSTM, and CNN2d (Table\ref{tab:baselines_10}) . These comparisons go beyond the strict parameter-matching paradigm outlined in the main paper, allowing architectural latitude in the baseline models to showcase their maximum potential performance under flexible conditions.

The motivations for this experiment are twofold. First, it addresses the practical concern that simpler models, such as MLPs or LSTMs, may perform comparably with smaller datasets. By demonstrating the superior performance of the transformer-based model even when baselines are given additional parameters, we underscore its robustness and suitability for complex layout tasks like knolling. Secondly, this comparison contributes to understanding the limitations of conventional architectures in capturing spatial dependencies critical to knolling applications.

Our transformer-based model significantly outperforms these baselines, where we report the error and parameter count for each model and configuration. This comparison illustrates that even with increased parameters, the baseline architectures exhibit higher error rates, reinforcing the advantages of a transformer-based approach for Knolling task.

\begin{table*}[!ht]
    \small
    \centering
    \caption{\label{tab:baselines_10}Comparison to more baselines}
    \begin{tabular}{ccccccccc}
        \textbf{MLP} & \textbf{Error} & \textbf{Parameters} & \textbf{LSTM} & \textbf{Error} & \textbf{Parameters} & \textbf{CNN2d} & \textbf{Error} & \textbf{Parameters } \\ 
        1 & 3.73E-01 & 62079 & 1 & 3.15E-02 & 183732 & 1 & 3.08E-02 & 92930  \\ 
        2 & 3.46E-01 & 107205 & 2 & 3.06E-02 & 101740 & 2 & 3.01E-01 & 92930  \\ 
        3 & 3.20E-01 & 10885 & 3 & 3.14E-02 & 73152 & 3 & 3.99E-01 & 40081  \\ 
        4 & 3.25E-01 & 125540 & 4 & 2.70E-02 & 191480 & 4 & 3.95E-01 & 42520  \\ 
        5 & 1.49E-01 & 101979 & 5 & 2.91E-02 & 103236 & 5 & 4.00E-01 & 36941  \\ 
        6 & 2.98E-01 & 136815 & 6 & 2.57E-02 & 60622 & 6 & 3.62E-01 & 30344  \\ 
        7 & 1.40E-01 & 83100 & 7 & 3.15E-02 & 62372 & 7 & 3.42E-01 & 79924  \\ 
        8 & 2.32E-01 & 8425 & 8 & 2.91E-02 & 81524 & 8 & 4.67E-01 & 27620  \\ 
        9 & 2.07E-01 & 70335 & 9 & 3.09E-02 & 90990 & 9 & 4.01E-01 & 34670  \\ 
        10 & 3.41E-01 & 170025 & 10 & 2.32E-02 & 183732 & 10 & 3.93E-01 & 33196  \\ 
        MEAN & 2.73E-01 & ~ & MEAN & 2.90E-02 & ~ & MEAN & 3.49E-01 &   \\ 
        STD & 8.02E-02 & ~ & STD & 2.69E-03 & ~ & STD & 1.14E-01 &   \\ 
        MIN & 1.40E-01 & ~ & MIN & 2.32E-02 & ~ & MIN & 3.08E-02 &   \\ 
    \end{tabular}
\end{table*}

\section{Quantitative Evaluation and Ablation Study}
To quantitatively assess the performance of our knolling model, we employed the L1 distance metric, which measures the absolute difference between the predicted and ground truth object positions. For deterministic results during evaluation, we selected the predicted Gaussian distribution based on minimum loss instead of the mixture weights. Our evaluation was conducted on a diverse test dataset comprising 20,000 samples for each object count ranging from 2 to 10 items, ensuring a comprehensive assessment across various knolling scenarios.

\begin{table}[h!]
    \caption{Comparison between Our Method (OM) and Baselines (LSTM, MLP)}
    \label{tab:sim_eval}
    \centering
    \scriptsize  
    \setlength{\tabcolsep}{3pt}  
    \begin{tabular}{ccccccc}
    \hline
        ~ & N objs. & 2 & 4 & 6 & 8 & 10  \\ \hline
        ~ & MEAN & 3.38E-04 & 2.40E-04 & 1.72E-04 & 2.08E-04 & 3.06E-04  \\ 
        OM & STD & 2.03E-04 & 2.71E-04 & 2.94E-04 & 4.13E-04 & 5.51E-04  \\ 
        ~ & MIN & 8.52E-05 & 2.60E-06 & 3.40E-07 & 3.88E-06 & 1.36E-05  \\ 
        ~ & MAX & 1.74E-03 & 2.16E-03 & 3.83E-03 & 4.95E-03 & 7.77E-03  \\ \hline
        ~ & MEAN & 1.72E-02 & 2.09E-02 & 2.35E-02 & 2.62E-02 & 2.98E-02  \\ 
        LSTM & STD & 3.78E-03 & 3.53E-03 & 4.15E-03 & 3.88E-03 & 6.36E-03  \\ 
        ~ & MIN & 8.37E-03 & 1.20E-02 & 1.43E-02 & 1.63E-02 & 1.86E-02  \\ 
        ~ & MAX & 2.65E-02 & 3.10E-02 & 3.47E-02 & 4.37E-02 & 4.99E-02  \\ \hline
        ~ & MEAN & 2.17E-01 & 1.45E-02 & 1.90E-01 & 2.55E-01 & 3.26E-01  \\ 
        MLP & STD & 1.91E-03 & 4.68E-03 & 2.42E-03 & 7.97E-03 & 8.95E-03  \\ 
        ~ & MIN & 2.14E-01 & 6.98E-03 & 1.81E-01 & 2.38E-01 & 3.07E-01  \\ 
        ~ & MAX & 2.23E-01 & 3.07E-02 & 2.00E-01 & 2.83E-01 & 3.57E-01  \\ \hline
    \end{tabular}
    \vspace{-10pt}
\end{table}

\begin{table*}
\caption{Ablation study}
\label{tab:ablation}
\resizebox{\textwidth}{!}{
\begin{tabular}{|c|l|c|c|c|c|c|}
\hline
Model     & \multicolumn{1}{c|}{Test Performace} & Log-likelihood Loss & MSE Loss  & MSE Min Loss & Overlapping Loss & Entropy Loss \\ \hline
OM & \textbf{5.78E-04 $\pm$ 6.64E-04} & -6.37E+00 $\pm$ 1.25E+00 & 4.91E-04 $\pm$ 6.25E-04 & 8.76E-03 $\pm$ 1.60E-02 & 9.23E-01 $\pm$ 9.55E-03 & 2.55E-02 $\pm$ 1.43E-02 \\
OM w/o LL. Loss & 7.24E-04 $\pm$ 7.53E-04 & 1.13E+02 $\pm$ 1.61E+02 & 6.34E-04 $\pm$ 7.05E-04 & 9.07E-03 $\pm$ 1.43E-02 & 9.01E-01 $\pm$ 2.25E-08 & 3.56E-02 $\pm$ 8.82E-03 \\
OM w/o Pos. Loss & 1.23E-02 $\pm$ 2.74E-03 & -1.50E+00 $\pm$ 4.53E-01 & 1.05E-02 $\pm$ 2.47E-03 & 1.80E-01 $\pm$ 1.25E-01 & 1.06E+00 $\pm$ 4.09E-03 & 1.71E-01 $\pm$ 1.20E-02 \\
OM w/o Pos. Min Loss & 3.93E-02 $\pm$ 7.48E-03 & 3.51E-01 $\pm$ 5.64E-01 & 3.90E-02 $\pm$ 7.48E-03 & 2.32E-02 $\pm$ 3.37E-02 & 1.09E+00 $\pm$ 4.73E-03 & 1.14E-01 $\pm$ 1.13E-02 \\
OM w/o Over. Loss & 5.08E-02 $\pm$ 1.22E-02 & -6.92E+00 $\pm$ 1.55E+00 & 3.40E-04 $\pm$ 4.07E-04 & 5.05E+00 $\pm$ 1.22E+00 & 9.66E-01 $\pm$ 2.44E-02 & 1.95E-02 $\pm$ 1.24E-02 \\
OM w/o Ent. Loss & 1.03E-03 $\pm$ 1.00E-03 & -4.71E+00 $\pm$ 1.04E+00 & 9.07E-04 $\pm$ 9.43E-04 & 1.20E-02 $\pm$ 1.97E-02 & 1.08E+00 $\pm$ 5.70E-02 & 2.54E-02 $\pm$ 1.27E-02 \\ \hline
\end{tabular}
}
\end{table*}

Two baseline models were established to enable a performance comparison to Our Method (OM). The first baseline model utilized a Multilayer Perceptron (MLP) architecture, and the second baseline was designed based on the Long Short-Term Memory (LSTM) model. We use LSTM as a baseline due to its effectiveness in sequence prediction tasks. Arranging objects can be framed as predicting a sequence of positions, where each object’s placement is influenced by prior placements. LSTMs have been extensively used for various sequence prediction tasks and have shown good performance. Thus, LSTMs serve as a more challenging baseline for benchmarking the performance of our transformer-based model. Moreover, comparing Transformer model with LSTMs helps to highlight the benefits of using self-attention and auto-regression in the transformer for handling variable input and output sizes and multi-label problems. Just as we did with Transformer model, we measured the L1 distance between the actual and predicted positions for each baseline model, with the results detailed in Table \ref{tab:sim_eval}. As a commitment to a fair evaluation, we have ensured that each model utilizes a similar amount of parameters: Transformer model incorporates 87,458 parameters, the LSTM baseline uses 86,858 parameters, and the MLP baseline operates with 87,788 parameters. Our knolling model consistently outperforms the MLP and LSTM baselines in terms of the L1 error. This superiority of Transformer model is evident in all parameters: mean L1 error, standard deviation, and minimum and maximum error. These results validate the effectiveness of our transformer-based model, which employs self-attention and auto-regression in handling varying input sizes and multi-label problems in knolling tasks.

Furthermore, we conducted an ablation study to evaluate the contributions of the individual loss components employed during training (Table \ref{tab:ablation}). The ablation results demonstrated the importance of each loss term, with their removal leading to performance degradation, justifying the need for a comprehensive loss function to capture the multifaceted nature of knolling tasks effectively.

\section{Transformer Performance across Dataset Sizes}
The conducted experiment provides valuable insights into how the performance of our knolling model scales with the size of the training dataset. As shown in Table \ref{tab:sup_experiment}, there is a clear and consistent decline in test error as the dataset size increases from 125,000 to 1 million samples. This reduction in error underscores the model's enhanced ability to generalize from larger amounts of data.

These results align with the well-established scaling laws in transformer architectures, where increasing the quantity of training data leads to improved model performance \cite{kaplan2020scaling}. The transformer-based knolling model benefits from exposure to a more extensive variety of object arrangements and spatial configurations, allowing it to learn more nuanced representations of tidiness and to better predict optimal object placements.

The implications of this experiment are significant for the future development of robotic tidiness tasks. The observed trend suggests that further increasing the dataset size could continue to enhance the model's performance. Moreover, by incorporating richer data representations, such as three-dimensional spatial information, the model could be extended to handle more complex environments.

This scaling behavior opens the possibility of generalizing our approach beyond tabletop organization to the tidying of entire rooms. With 3D data, the model could learn to navigate and organize objects within a full spatial context, considering factors like object height, room layout, and furniture placement. This would enable robots to perform comprehensive housekeeping tasks, such as arranging furniture, organizing shelves, and decluttering living spaces, ultimately moving closer to autonomous robots that can assist with daily household chores. To support reproducibility and transparency, we note that all training runs were conducted on a single NVIDIA RTX 3090 GPU. Training on 1 million samples required approximately 48 hours.

\begin{table}[!ht]
    \centering
    \caption{Test Error vs Dataset Size}
    \label{tab:sup_experiment}
    \begin{tabular}{ccccc}
    \hline
        \textbf{} & \textbf{Mean} & \textbf{Std} & \textbf{Min} & \textbf{Max } \\ \hline
        125K & 6.46E-04 & 1.82E-03 & 1.39E-15 & 3.17E-02  \\ 
        250K & 5.81E-04 & 1.81E-03 & 0.00E+00 & 3.21E-02  \\ 
        500K & 5.18E-04 & 1.79E-03 & 1.39E-15 & 3.53E-02  \\ 
        1M & 4.29E-04 & 1.61E-03 & 0.00E+00 & 3.49E-02  \\ 
        1M-Fine & 3.57E-04 & 1.54E-03 & 0.00E+00 & 3.44E-02  \\ \hline
    \end{tabular}
    \vspace{-3pt}
\end{table}

\section{Knolling Pipeline And Training Objective Details}
The visual perception model is a customized YOLO v8 model \cite{reis2023real}, fine-tuned to predict the information of the objects, including the segmentation, color, and category. We post-process the segmentation to a rectangular box and then extract the object state, including the center position, orientation, and dimensions. Detailed evaluation results and confusion matrices for the visual perception model in both simulated and real-world environments are provided in the supplementary materials. The data required for training our models is generated within the Pybullet simulation environment for pre-training and collected in the real world for fine-tuning. To prepare our Visual Perception Model for real-world deployment and address the simulation-to-reality gap, we apply the visual domain randomization technique to our data collection process \cite{tremblay2018training}. This technique helps in enhancing the model's adaptability and robustness by introducing variations and uncertainties that mimic real-world conditions. We manipulate several factors within the simulation environment, such as brightness and ground textures. Using this approach, we train our model to disregard extraneous noise and focus on essential features.

The robotic arm controller operates in four modes: movement, grasping/releasing, table sweeping, and object separation. It uses the predicted object positions from the knolling model and the current poses from the perception module to execute smooth pick-and-place operations, clearing occupied spaces through sweeping motions when necessary.

This integrated pipeline demonstrates how our self-supervised knolling model can be leveraged for practical robotic organization tasks, enabling autonomous tidying of cluttered environments without explicit human supervision or target specifications.

\subsection{Training Objectives}
Our training strategy incorporates multiple loss functions to guide the model toward accurate, diverse, and physically feasible object placements. These include log-likelihood loss to capture multimodality, MSE-based losses to encourage precision, and specialized penalties to discourage overlap and encourage solution diversity. The specific formulation of each loss component, including detailed equations and justifications, is provided in the Supplementary Material.

\section{Training Pipeline Details}

\textbf{Pre-training with Self-supervised Learning.} To enable the model to learn underlying structural relationships in object arrangements without explicit supervision, we employ a self-supervised pre-training strategy. In this phase, we mask partial object locations and train the model to predict the next object's position based solely on the existing object states. This approach is analogous to masked language modeling in NLP, where models learn to predict missing words in a sentence \cite{devlin2018bert}. By training the model to predict masked object positions, it learns the spatial dependencies and organizational patterns inherent in tidy arrangements. 

In the pre-training phase, the model is exposed to partially organized scenes, where it needs to predict the placement of the next object given the current arrangement. Formally, let $O = {o_1, o_2, \ldots, o_N}$ be the set of $N$ objects, and $P = {p_1, p_2, \ldots, p_{N-1}}$ be the given positions of the first $N-1$ objects. The pre-training objective is to predict the position $p_N$ of the remaining object $o_N$:

\begin{equation} p_N = f_{\text{pre-train}}(O, P) \end{equation}

This pre-training stage allows the model to learn the fundamental principles of spatial organization and object placement without the added complexity of predicting all object positions simultaneously. By masking the position of the next object and training the model to predict it based on the current arrangement, the model learns underlying patterns and relationships in object arrangements without explicit human annotations.

\textbf{Fine-tuning the Knolling Model.}
The fine-tuning phase is dedicated to refining the model's ability to execute complete knolling tasks from scratch, starting with cluttered or disorganized scenes. In this stage, the model is tasked with predicting the positions $P = {p_1, p_2, \ldots, p_N}$ of all $N$ objects given only their dimensions $O = {o_1, o_2, \ldots, o_N}$:

\begin{equation} P = f_{\text{fine-tune}}(O), O = \{o_n | o_n = (w_n, l_n), n \in \mathbb{N}^+\}. \end{equation}

During fine-tuning, the model leverages the knowledge acquired during pre-training to predict the positions of all objects in an autoregressive manner. At each step, the model predicts the position of the next object based on the previously predicted positions and the objects' dimensions. This process can be formalized as:

\begin{equation} p_n = f_{\text{fine-tune}}(O, {p_1, p_2, \ldots, p_{n-1}}), \quad n = 1, 2, \ldots, N \end{equation}

By progressively predicting the positions of all objects, the model refines its understanding of complete tidy arrangements, enabling it to handle arbitrary numbers of objects and diverse configurations. The details about dataset are shown in the supplementary materials. The training process leverages a combination of five loss functions to train a model to predict object positions without overlaps while also promoting diversity in predictions.